\documentclass[runningheads]{llncs}

\usepackage{eccv}

\usepackage{eccvabbrv}

\usepackage{graphicx}
\usepackage{booktabs}

\usepackage{tabularx}
\usepackage{multirow}
\usepackage[table]{xcolor}
\usepackage{amssymb}

\usepackage[accsupp]{axessibility}  

\usepackage{hyperref}

\usepackage{orcidlink}

\begin{document}

\title{MUSE: Unlocking Timestep as Native Task Steering for One-Step Dense Prediction} 

\titlerunning{Unlocking Timestep as Native Task Steering for One-Step Dense Prediction}

\author{Shuo Zhou\inst{1,2}\orcidlink{0000-0002-0108-8222} \and
Zhaoxin Li\inst{1}\orcidlink{0000-0001-5450-8131} \and
Xiujuan Chai\inst{2}\orcidlink{0000-0002-2757-9900}\thanks{Corresponding author.}}

\authorrunning{S. Zhou et al.}

\institute{Agricultural Information Institute, Chinese Academy of Agricultural Sciences, Beijing, China.
\email{\{zhoushuo,lizhaoxin\}@caas.cn}
\and
Key Laboratory of Agricultural Big Data, Ministry of Agriculture and Rural Affairs, Beijing, China.
\email{chaixiujuan@caas.cn}}

\maketitle

\begin{abstract}
  Monocular dense prediction has recently seen remarkable success by repurposing pre-trained diffusion models.
  This opens a promising yet challenging avenue for more efficient multi-task learning paradigm.
  However, existing multi-task diffusion methods often introduce parameter-heavy adapters,
   experts, or learnable task tokens, leading to computational redundancy.
  In this paper, we reveal an inherent mechanism within one-step diffusion models:
   the native, fixed timestep positional embedding can be repurposed as an endogenous task steering signal.
  Based on this discovery, we propose Multi-task Unified eStimation via timestep Embedding (MUSE),
   a parameter-free, single-model multi-tasking approach for dense prediction.
  We interpret this mechanism via Manifold Decoupling, where
   discrete, fixed timestep values deterministically steer the generation process towards
   decoupled, task-specific manifolds in the latent space.
  Extensive experiments across 10 datasets demonstrate that 
   MUSE achieves highly competitive performance on both monocular depth and normal estimation,
   and its efficacy generalizes across U-Net and DiT architectures.
  Our work offers a concise and efficient path toward generalist vision models
   by simply unlocking the latent potential of existing generation infrastructure.
\end{abstract}

\begin{figure}[ht]
  \centering
  \includegraphics[width=0.96\textwidth]{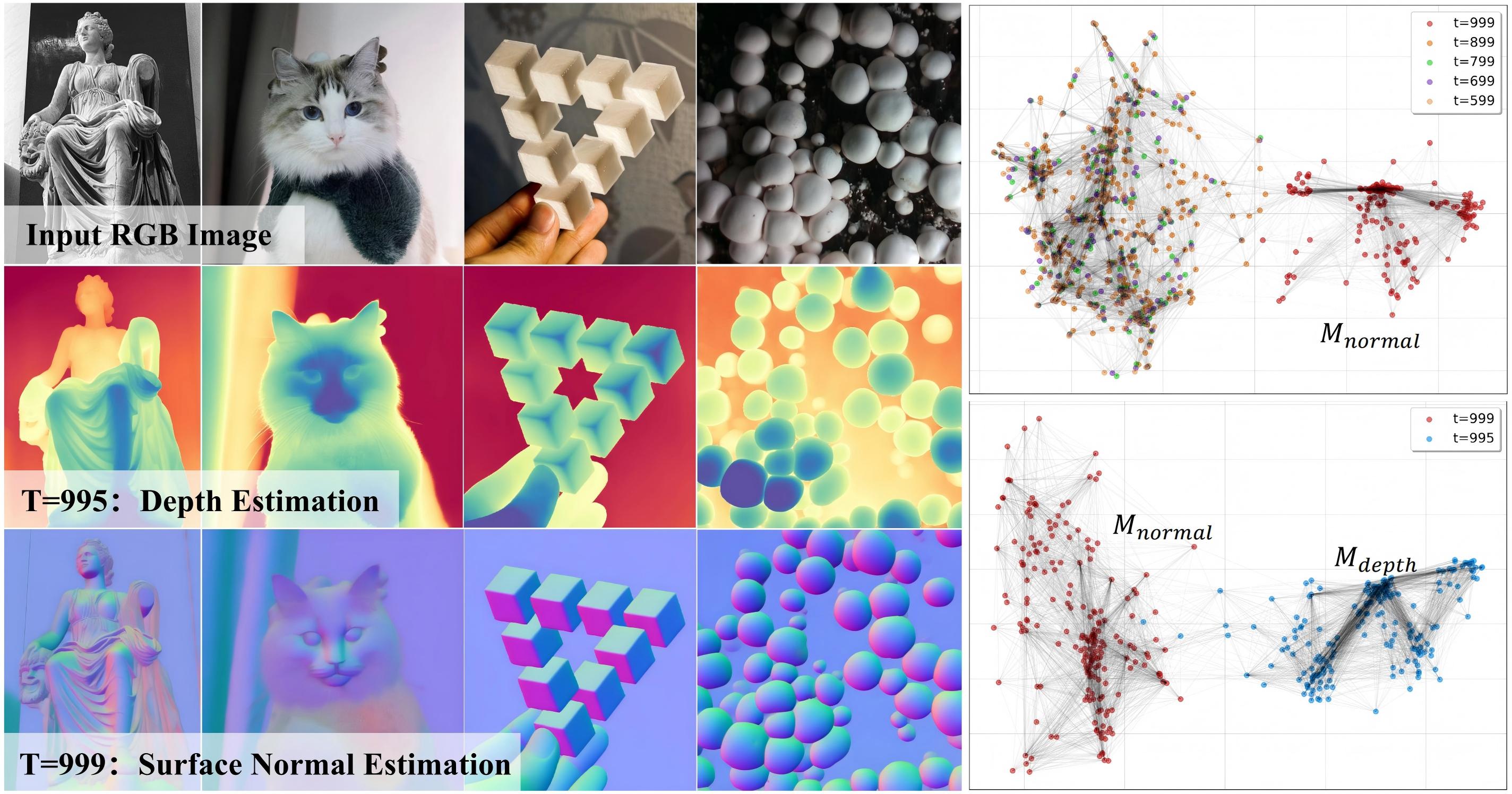}
  \caption{We present MUSE, a single-model multi-tasking framework based on one-step diffusion for monocular dense prediction.
    Left: Qualitative overview of zero-shot depth and normal estimation by MUSE in challenging and ambiguous scenes.
    Right: Isomap 2D visualization of sample's latent features produced by a MUSE model which is steered with timestep $t_{\text{depth}}$=995 and $t_{\text{normal}}$=999.
    Features conditioned by steering timesteps are decoupled to distinct manifolds and exhibit clear separation,
     whereas those without being steered remain entangled, proving the Manifold Decoupling property, see Sec \ref{sec:manifold} for more details.
  }
  \label{fig:teaser}
\end{figure}

\section{Introduction}
\label{sec:intro}

Geometric estimation tasks form the bedrock of spatial intelligence for autonomous agents.
A machine's ability to perceive and reason about the 3D world from a single RGB image is critical for its large-scale applications.
In recent years, the field has witnessed a paradigm shift with the advent of
 text-to-image (T2I) diffusion models \cite{ho2020denoising, song2020denoising, rombach2022high, lai2025principles}.
Pre-trained on web-scale datasets \cite{schuhmann2022laion}, these models have developed an encyclopedic prior of the natural world.
Consequently, a burgeoning branch of research has focused on \textit{Repurposing} these powerful priors for visual perceiving tasks.
Since the pioneering research of \cite{saxena2023monocular, ji2023ddp, duan2024diffusiondepth},
 taming conditional latent diffusion models (LDMs) \cite{rombach2022high}
 for monocular depth and normal estimation has been extensively explored \cite{ke2024repurposing, he2024lotus, krishnan2025orchid, he2025lotus, wang2025jasmine}.
The rich knowledge learned by the pre-trained VAE enables them to achieve competitive zero-shot performance
 by fine-tuning on a small number of high-quality synthetic data, demonstrating remarkable data efficiency and generalization capabilities.

Naturally, the success in single-task scenarios motivates the next frontier: 
 developing a unified model for diverse dense prediction.
However, extending diffusion to a multi-task setting is non-trivial.
Prevailing approaches often resort to explicit architectural modifications,
 such as multi-head decoders \cite{ye2024diffusionmtl}, parameter-heavy Mixture-of-Experts
  (MoE) \cite{dong2024unidense}, or the injection of additional learnable task tokens and adapters \cite{zhang2024three, long2024wonder3d, fu2024geowizard}.
Furthermore, attempting to jointly optimize a single shared network for disparate tasks
 often leads to gradient conflicts and negative transfer \cite{go2023addressing, hang2023efficient, javaloy2021rotograd}.
This raises a fundamental question: 
 Must we introduce new parameters and complex modules to build a unified perception model,
 or does the pre-trained diffusion infrastructure already possess the latent capacity to decouple multiple tasks?

In this paper, we shift the focus from designing external modules to mechanism discovery.
We present a counter-intuitive but remarkably effective insight:
 within a one-step diffusion framework, the native, fixed sinusoidal timestep embedding — originally designed solely to indicate noise levels —
 can be repurposed as an endogenous and parameter-free task steering signal. 
Based on this, we propose the Multi-task Unified eStimation via timestep Embedding (MUSE).
MUSE operates under a single-model multi-tasking paradigm in a truly minimalist fashion:
 by simply assigning discrete, constant timestep values to different tasks during fine-tuning and inference,
 it conditions a single, fully parameter-shared U-Net or DiT \cite{peebles2023scalable} to activate corresponding task-specific capabilities.
MUSE achieves this without introducing any extra learnable task tokens, specialized loss functions,
 or considerable structural changes compared to the vanilla diffusion-based dense prediction models.
Moreover, to elucidate the underlying mechanics, we provide a geometric interpretation termed \textit{Manifold Decoupling}.
By assigning orthogonal keys to each task,
 their respective optimization trajectories are fundamentally decoupled in the latent geometric space,
 elegantly avoiding the gradient conflicts that cause negative transfer.
In this paper,``orthogonal'' denotes functional disjointness and subspace decoupling, not zero dot-products.

Our main contributions are summarized as follows:
\begin{itemize}
  \item For the first time we reveal a novel endogenous mechanism in one-step diffusion paradigm:
   the native, fixed timestep can function as a parameter-free semantic switch to steer distinct prediction tasks.
  \item MUSE, an elegant fine-tuning and inference protocol for unified monocular depth and normal estimation is proposed.
   Extensive experiments on 10 diverse benchmarks demonstrate that it achieves highly competitive performance,
   with generality across both U-Net and DiT architectures.
  \item We theorize and validate this mechanism through Manifold Decoupling,
   providing extensive geometric visualizations of the latent space separation.
\end{itemize}

\section{Related Work}
\subsection{LDMs for Monocular Dense Prediction}

The remarkable success of pre-trained T2I diffusion models has inspired a new research paradigm: 
 repurposing their powerful visual priors for monocular dense prediction
  \cite{saxena2023monocular, duan2024diffusiondepth, long2024wonder3d, ji2023ddp, ke2024repurposing, he2024lotus, fu2024geowizard, lee2024exploiting, xu2024matters, zhang2024three, song2025depthmaster, krishnan2025orchid}.

\textbf{Multi-Step Diffusion.} 
Early milestone works \cite{saxena2023monocular, ke2024repurposing} formulated this task as a conditional image generation problem. 
They operated under a stochastic, multi-step paradigm, where a target prediction map is iteratively denoised from random Gaussian noise, 
 conditioned on the input RGB image. 
While effective, these methods are computationally intensive and suffer from inherent randomness, 
 often requiring test-time ensembling \cite{ke2025marigold} to produce stable results.

\textbf{Single-Step Diffusion.}
Subsequent research has identified that the deterministic nature of dense prediction tasks
 is misaligned with the stochastic, noise-to-data probabilistic generation process. 
DMP \cite{lee2024exploiting} reformulated the diffusion process as a discriminative mapping
 between the image and the prediction by eliminating the random noise input.
Works like GenPercept \cite{xu2024matters}, Lotus \cite{he2024lotus,he2025lotus}, and DepthMaster \cite{song2025depthmaster}
 demonstrated that the multi-step iterative process is not essential for better performance,
 and a one-step, noise-free prediction can achieve or even surpass the performance of multi-step methods, 
 while being orders of magnitude faster. 

Along with some most recent explorations of optimizing one-step generation models \cite{frans2024one, geng2025mean, chen2025steering, you2025modular},
 a crucial question arises: if the iterative process is unnecessary,
 what role does the timestep play beyond its traditional function as a noise level indicator?
By peeling back the mathematical abstraction and delving into the implementation details,
 we unlock the potential of the timestep embedding as a powerful, yet underexplored, conditioning signal.

\subsection{LDMs for Multi-Task Learning}

The application of diffusion models to multi-task learning (MTL) can be broadly viewed from two perspectives.

\textbf{Explicit MTL Architectures.}
One line of work focused on designing explicit architectures to enable a single diffusion model to handle multiple tasks. 
Some adopted the commonly used One-backbone with Multi-head design to generate outputs for different tasks \cite{ye2024diffusionmtl}. 
Others \cite{dong2024unidense, park2024switch} transformed the U-Net into a Mixture-of-Experts (MoE) architecture 
 to select task-specific computational paths. 
Orchid \cite{krishnan2025orchid} and TaskDiffusion \cite{yangmulti} proposed joint learning schemes by unifying task labels into a shared representation space,
 where the former even retrained the VAE module in a joint color-geometry latent space.
Diception \cite{zhao2025diception} adapted a Diffusion Transformer via lightweight task embeddings and point prompt, achieving strong multi-task capability.
A more detailed breakdown of the diffusion model's conditional mechanisms can be found in the Supp. Materials Sec.A.

\textbf{Implicit MTL Strategy.} 
Another insightful perspective rethought the standard training of a diffusion model as an implicit form of MTL, 
 where each timestep represents a distinct denoising task \cite{go2023addressing, hang2023efficient, ma2025decouple}. 
This viewpoint disclosed a fundamental challenge: the optimization objectives for different timesteps, 
 especially those with disparate noise levels, can conflict.
This results in conflicting gradients and leads to negative transfer \cite{crawshaw2020multi},
 a classic problem in MTL that impedes training convergence and degrades task performance.
To mitigate this, ANT \cite{go2023addressing} proposed clustering timesteps and applying MTL optimization techniques at the cluster level,
 Min-SNR \cite{hang2023efficient} introduced a weighting scheme to balance the contributions of different timesteps,
 while DeMe \cite{ma2025decouple} decoupled the training by fine-tuning separate models for different timestep ranges and then merging them.
The accessibility of these methods is limited by their reliance on massive computing power, complex and cumbersome models, or extremely large trainset.

\subsection{LDMs and Manifold Learning}
The manifold hypothesis \cite{seung2000manifold} posits that high-dimensional data lies on a low-dimensional manifold,
 revealing how a model captures the intricate structure of data distributions.
Several studies have leveraged this geometric perspective to analyze diffusion models. 
Chung et al. \cite{chung2022improving} enforced manifold constraints by 
 projecting the intermediate results back onto a manifold defined by physical constraints at each denoising step. 
Similarly, He et al. \cite{he2023manifold} preserved the manifold structure during guided diffusion by 
 projecting the guidance signal onto the manifold's tangent space. 
Stanczuk et al. \cite{stanczuk2022your} compellingly showed that 
 a trained diffusion model implicitly learns the intrinsic dimension of the data manifold, 
 which can be estimated from the Jacobian of the score function. 
Jacobsen et al. \cite{jacobsen2025staying} proposed 
 injecting noise along the tangent space of the data manifold to better preserve its structure.
Recently, the JiT \cite{li2025back} has brought manifold learning to the forefront of the research community's attention.
These works provide the theoretical and mathematical foundation for our Manifold Decoupling interpretation.

\section{Method}
\label{sec:method}

\subsection{Preliminaries}
\textbf{Classical Denoising Diffusion Probabilistic Models.} 
A standard DDPM \cite{ho2020denoising} consists of a forward noising process and a reverse denoising process. 
The forward process gradually perturbs clean data $x_0$ over $T$ steps 
 by adding Gaussian noise according to a predefined variance schedule $\{\beta_t\}_{t=1}^T$:
\begin{equation}
  \label{eq:ddpm}
  q(\mathbf{x}_t | \mathbf{x}_0) = \mathcal{N}(\mathbf{x}_t; \sqrt{\bar{\alpha}_t}\mathbf{x}_0, (1-\bar{\alpha}_t)\mathbf{I})
\end{equation}
where $\alpha_t = 1 - \beta_t$ and $\bar{\alpha}_t = \prod_{s=1}^t \alpha_s$. 
The reverse process aims to recover the clean data from pure Gaussian noise $\mathbf{x}_T \sim \mathcal{N}(0, \mathbf{I})$
by learning a neural network $\epsilon_\theta(\mathbf{x}_t, t)$ to predict the added noise $\epsilon$ at each timestep $t$ from the noisy input $\mathbf{x}_t$.
The model is trained by optimizing a simplified objective:
\begin{equation}
  \label{eq:ddpm_objective}
  \mathcal{L}_{\text{DDPM}} = \mathbb{E}_{t, \mathbf{x}_0, \epsilon} \left[ || \epsilon - \epsilon_\theta(\sqrt{\bar{\alpha}_t}\mathbf{x}_0 + \sqrt{1-\bar{\alpha}_t}\epsilon, t) ||^2 \right]
\end{equation}
Inference is an iterative process that starts from $\mathbf{x}_T$ and denoises it step-by-step to obtain the final sample $\mathbf{x}_0$.

\textbf{Conditional Diffusion for Dense Prediction.} 
Modern approaches like Stable Diffusion (SD) \cite{rombach2022high} operate in a compressed latent space. 
It is provided by a pre-trained Variational Autoencoder (VAE) with a pair of encoder $\mathcal{E}$ and a decoder $\mathcal{D}$.
For a dense prediction task, given an input RGB image $I$ and its corresponding ground-truth map $Y$,
 their latent representations are obtained as $\mathbf{z}^{x} = \mathcal{E}(I)$ and $\mathbf{z}^{y} = \mathcal{E}(Y)$, respectively.
The task is then formulated as learning a conditional denoiser $\epsilon_\theta(z_t, t, c)$ that predicts the noise from the noisy target latent $z_t$, conditioned on $c$.
For dense prediction, $c$ is the input image latent feature, $c = \mathbf{z}^{x}$.

\textbf{One-Step and Deterministic Paradigm.} 
Recent advancements, epitomized by Lotus \cite{he2024lotus}, 
 have demonstrated that for discriminative dense prediction tasks, the iterative multi-step denoising process is not essential. 
Instead, a more efficient single-step paradigm can be adopted, which simplifies the process in two key ways. 
First, it opts for directly predicting the clean latent $\mathbf{z}^{y}$ (termed $x_0$-prediction) rather than the noise $\epsilon$ ($\epsilon$-prediction).
Second, it collapses the training and inference into a single, fixed timestep, typically the maximum timestep $t=T$ ($T=999$ in SD).
In the deterministic variant, the random noise component is entirely removed from the input,
 and surprisingly achieves the best performance on both depth and nomral estimation tasks.
The model, now denoted as $f_\theta$,
 learns a direct mapping from the conditional image latent $\mathbf{z}^{x}$ to the target prediction latent $\mathbf{z}^{y}$.
The training objective simplifies to a direct regression:
\begin{equation}
  \label{eq:lotu}
  \mathcal{L}_{\text{one-step}} = \mathbb{E}_{\mathbf{I}, \mathbf{Y}} \left[ || \mathbf{z}^{y} - f_\theta(\mathbf{z}^{x}, T) ||^2 \right]
\end{equation}
where $T$ is the fixed timestep embedding. 

\subsection{MUSE: Multi-task Unified Estimation via Timestep Embedding}

The timestep embedding, once a dynamic indicator of noise levels, becomes a static and underutilized signal in one-step generation. 
This paper posits that this native and powerful conditioning channel can be repurposed as a semantic switch for multi-task learning. 

\subsubsection{Repurposing Timestep as the Task Steering}

We propose that a single SD U-Net can be taught to perform distinct tasks by associating each task with a unique, discrete timestep value. 
Instead of treating the timestep $t$ as a continuous variable representing noise levels, we re-contextualize it as a categorical task prompt. 
This allows us to decouple the learning processes of multiple tasks in a parameter-free and distillation-free manner.

Formally, we define a set of target tasks $\mathcal{S}=\{\text{Task}_1, ..., \text{Task}_K\}$,
 and pre-assign a unique timestep to each: $\mathcal{\tau}_{\text{task}} \in \{t_1, t_2, ..., t_k\}$.
In MUSE, we set the timestep values as $t_{\text{rgb}}$, $t_{\text{depth}}$, and $t_{\text{normal}}$
 corresponding to the tasks of RGB reconstruction, affine-invariant depth estimation and normal estimation.
The one-step prediction model $f_\theta$ is now conditioned on these task-specific timesteps:
\begin{equation}
  \label{eq:muse}
  \mathbf{z}_{\text{pred}}^{y} = f_\theta(\mathbf{z}^{x}, t_{\text{task}})
\end{equation}

This formulation elegantly transforms the single-task model into a multi-task learner without adding any new parameters or modules. 
The model learns to interpret the fixed embedding of these discrete timestep values as 
 instructions to activate different computational pathways within its vast parameter space, each specialized for a particular task.
The unified training objective for MUSE is a simple summation of the Mean Squared Error (MSE) loss for each task. 
For a given training sample ($I$, $Y_{\text{task}}$), where $task \in \mathcal{S}$, the loss is computed as:
\begin{equation}
  \label{eq:muse_loss}
  \mathcal{L}_{\text{MUSE}} = \mathbb{E}_{(\mathbf{I}, \mathbf{Y}_{\text{task}}), t_{\text{task}}} \left[ || \mathcal{E}(\mathbf{Y}_{\text{task}}) - f_\theta(\mathcal{E}(\mathbf{I}), t_{\text{task}}) ||^2 \right]
\end{equation}

MUSE requires no complex task-specific loss functions or sophisticated loss-balancing strategies. 
Basically, the decoupling is not achieved at the loss level, but at the input level, 
by providing orthogonal task steering ($t_{\text{k-1}} \neq t_{\text{k}}$) that guide the model to learn non-conflicting gradient paths for each task.

\begin{figure}[h]
    \centering
    \includegraphics[width=0.96\textwidth]{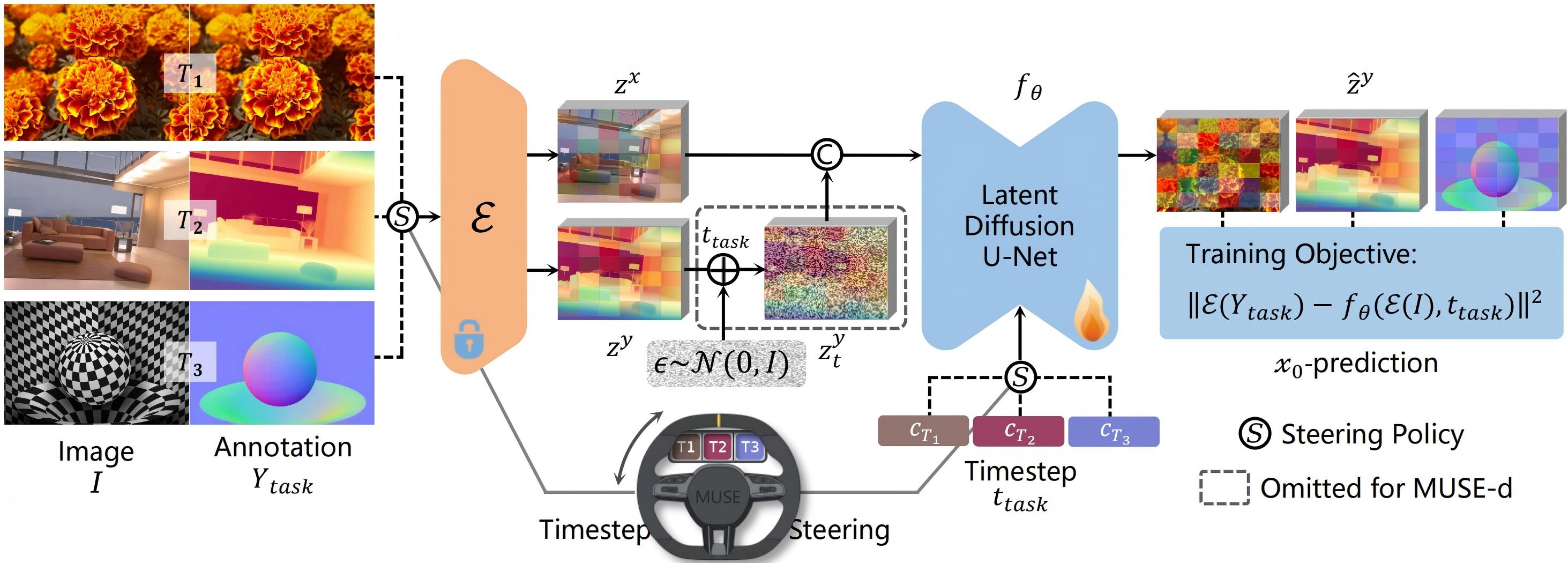}
    \caption{Overview of MUSE finetuning protocol. In training stage, multiple fixed timesteps are sampled by a steering policy to condition the loss calculation. 
    During inference, we can naturally use the timestep as a task steering to generate the corresponding prediction.
    By controlling the presence or absence of noise, generative or discriminative variant can be chosen.}
    \label{fig:muse}
\end{figure}

\subsubsection{Training and Inference Protocol}

The core of MUSE is the steering policy,
 involving a control of the data-loading and inference logic, as illustrated in Figure \ref{fig:muse}.

\textbf{Training.} 
Before all, the VAE of the pre-trained SD model is frozen, and only the U-Net parameters are fine-tuned.  
For each training iteration, we sample a mini-batch of training data as usual. 
While for every sample within the batch, the steering policy randomly endows it with one of the 3 task purposes:
 RGB reconstruction, depth estimation, or surface normal estimation. 
Then the task's pre-assigned timestep embedding is fed into the model along with the image latent feature.
The model's output is then computed only against the corresponding ground-truth latent of the selected task. 
The total loss for the batch is the average of these individual sample MSE losses,
 and the gradients are backpropagated to update the single, shared U-Net parameters.
In the baseline steering policy, the training frequency and loss contributions of sampled tasks is identical.
It can also be adjusted by task ratio and loss weight based on the difficulty of the tasks and the characteristics of the datasets.
This process forces the model to associate each specific timestep embedding with the goal of generating the corresponding task output.

\textbf{Inference.}
The inference process fully showcases the efficiency and elegance of MUSE.
To acquire predictions for all three tasks of a given input RGB image $I$,
 MUSE stacks the $I$ three times, assigns each $I$ a corresponding timestep,
 and then simultaneously outputs all the prediction maps in a single forward inference, thanks to the broadcast mechanism.
If specific scenarios and computational resource limitations are taken into account, MUSE can also be used as a single-task model.
For example, the model can be fed the $I$ along with timestep $t_{\text{depth}}$ for depth estimation task.
The output latent is decoded to produce the depth map $\hat{Y}_{\text{depth}} = \mathcal{D}(f_\theta(\mathcal{E}(I), t_{\text{depth}}))$.
The entire procedure is one-step for each task, requiring no iterative denoising. 
The ability to switch between tasks by simply changing a single integer input makes MUSE
 an extremely efficient and flexible framework for unified visual perception.
Some zero-shot depth and normal estimation are shown in Figure \ref{fig:teaser} and \ref{fig:timestep}.

\subsection{Formulation via Manifold Decoupling}

The Manifold Hypothesis \cite{stanczuk2022your, stanczuk2024diffusion, meilua2024manifold} states that high-dimensional real-world data, such as images of depth or surface normal, 
 do not populate the entire ambient space but rather concentrate on or near a low-dimensional manifold embedded within it. 
Inspired by these principles, we contend that MUSE's success is rooted in the geometric structure of the data it learns to encode in latent space.

Let $\mathcal{Z}$ be the high-dimensional latent space of the VAE. 
The latent codes of all valid depth maps lie on a low-dimensional manifold $\mathcal{M}_{\text{depth}} \subset \mathcal{Z}$,
 and similarly, normal data lie on another distinct $\mathcal{M}_{\text{normal}} \subset \mathcal{Z}$.
In a traditional multi-task learning scenario where parameters are shared through hard-coded methods, 
 the model must learn to generate outputs on both manifolds simultaneously. 
This often leads to conflicting gradients: an update that moves a prediction closer to $\mathcal{M}_{\text{depth}}$
 may inadvertently push it away from $\mathcal{M}_{\text{normal}}$, resulting in negative transfer.

The timestep embedding in MUSE acts as a geometric controllers that steers denoising dynamics along functionally segregated pathways. 
It conditions the function $f_\theta$ to learn a mapping that projects the image latent feature $\mathbf{z}^{x}$ specifically onto the manifold corresponding to that task. 
From this perspective, MUSE is not a single function, but a parameterized family of functions $\{f_{\theta,t}\}_{t \in \{t_{\text{task}}\}}$.
Each $f_{\theta,t_{\text{task}}}$ is trained to approximate a projection onto the target task manifold:
\begin{equation}
  \label{eq:manifold_projection}
  f_\theta(\mathbf{z}^{x}, t_{\text{task}}) \approx \Pi_{\mathcal{M}_{\text{task}}}(\mathbf{z}^{x})
\end{equation}

Any update to a point $z_{\text{pred}} \in \mathcal{M}_{\text{task}}$
 should ideally lie within the tangent space to ensure the updated point remains on the manifold \cite{he2023manifold}.
Because the model learns to associate the orthogonal inputs $t_{\text{task}}$ with updates along their respective, 
 and likely different manifolds, the optimization processes for different tasks do not interfere. 
MUSE learns distinct geometric objectives, guided by multiple distinct keys through the cross attention structure.

\section{Experiments and Results}
\label{sec:exp}
\subsection{Implementation}

We adopt the pre-trained weights of Stable Diffusion v2~\cite{rombach2022high} as the model initialization. 
All models are fine-tuned using the AdamW optimizer with a learning rate of $3 \times 10^{-5}$ and a total batch size of $16$. 
The models are trained for $12$K iterations on a single NVIDIA V100-32GB GPU. 
Following prevailing general pipelines \cite{ke2025marigold,he2024lotus}, we use a mixture of $9:1$ high-quality synthetic datasets, Hypersim~\cite{roberts2021hypersim} and Virtual KITTI~\cite{cabon2020vkitti2}, for fine-tuning.

We mainly implement two variants.
The generative paradigm MUSE-g follows the probabilistic diffusion setup,
 where a Gaussian noise is concatenated with the image latent $\mathbf{z}^{x}$ as input to the U-Net.
The discrimative paradigm MUSE-d is deterministic and removes the explicit noise input,
 and the U-Net only receives the image latent $\mathbf{z}^{x}$ and the task timestep embedding $c_{\text{time}}$.

We validate our approach on 10 acknowledged dense prediction benchmarks.
For affine-invariant depth estimation, we evaluate on NYUv2~\cite{silberman2012indoor},
 KITTI~\cite{geiger2013vision}, ScanNet~\cite{dai2017scannet}, ETD3D~\cite{schops2017multi}, and DIODE~\cite{vasiljevic2019diode}. 
For surface normal estimation, we evaluate on the NYUv2, ScanNet, iBims-1~\cite{koch2018evaluation}, Sintel~\cite{butler2012naturalistic}, and Oasis~\cite{chen2020oasis} datasets.
As widely recognized by the community, the reported performance metrics include: Absolute Relative Error (AbsRel$\downarrow$) and $\delta_1\uparrow$ accuracy for depth estimation,
 and Mean Angular Error (Mean$\downarrow$) and the percentage of pixels with an error below $11.25^{\circ}\uparrow$ for surface normal estimation.

\subsection{Mechanistic Analysis of Timestep-Driven Steering}
To understand exactly how the scalar timestep governs the shared U-Net, we conduct controlled ablations to dissect the underlying mechanism.
The design space of steering policy includes five dimensions: three timestep values corresponding to the three tasks, the frequency of task training, and the weights for loss calculation, as Table \ref{tab:setting}.  
We aim to answer two critical questions: Is the numerical distinctness of the timestep the sole driver of task decoupling, and does its absolute magnitude matter?

\begin{figure}[h]
  \centering
  \includegraphics[width=1.0\textwidth]{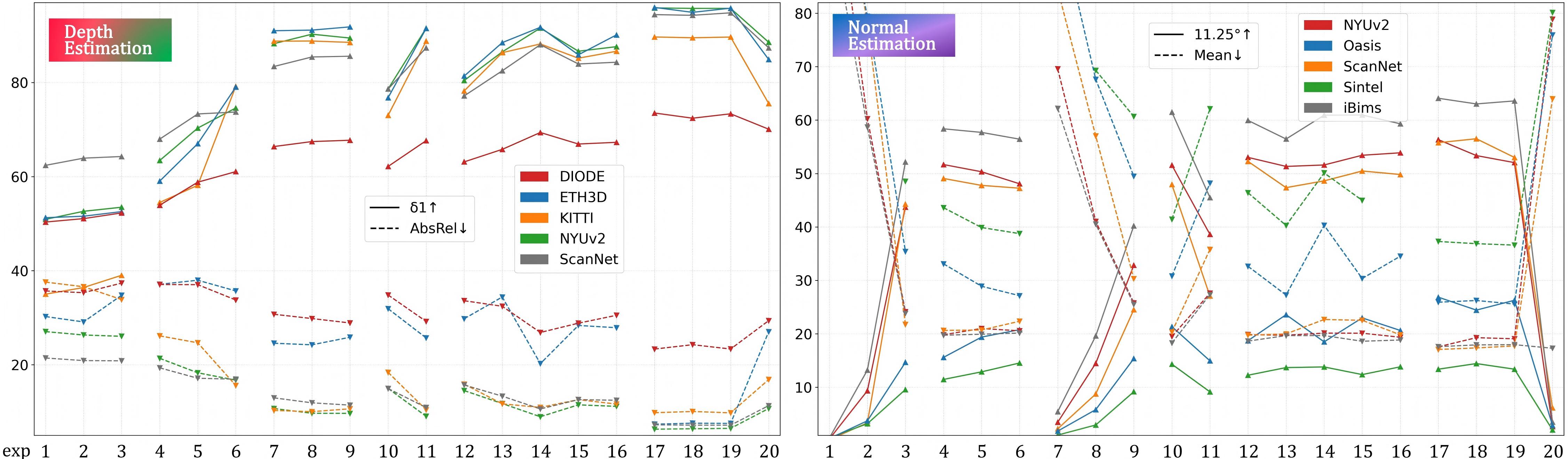}
  \caption{MUSE steer policy ablation study evaluation metrics.
     The experiment (\textit{exp}) numbers and groups of the horizontal axis correspond one-to-one with the 20 configurations in Table 1.
     The two metrics for each task are converted to [0,100] so they can be plotted in the same graph.
     Note that for every \textit{exp}, there are 10 metrics for depth estimation in the left diagram,
      plus another 10 metrics for normal estimation in the right diagram.
     See Supp. Materials Sec.B for complete numerical data.
  }
  \label{fig:ablation}
\end{figure}

\textbf{The Necessity of Orthogonal Steering Policy.}
As Figure \ref{fig:ablation} shows, we first establish the boundary conditions of negative transfer.
When the model is forced to predict both depth and surface normals using an identical timestep ($t_{depth} = t_{normal}$),
 it suffers from severe gradient conflicts and catastrophic interference, failing to converge on either task. 
However, the exact same architecture achieves simultaneous convergence simply by assigning discrete, distinct values ($t_{depth} \neq t_{normal}$). 
This empirically validates our core premise: the model does not require extra parameters;
 it relies entirely on the orthogonality of the given timestep embeddings to steer input features to their corresponding, decoupled task manifolds.

\begin{table}[h]
  \centering
  \caption{MUSE steer policy ablation study settings. 
    In the last line, ``tr'' represents adjusting the training frequency (task ratio)
     of RGB reconstruction, depth estimation, and normal estimation tasks to 1:2.5:2.5.
    ``lw'' indicates that the weights of the three tasks (loss weight) are adjusted to 1:2.5:2.5 when calculating the loss.
  }
  \label{tab:setting}
  \resizebox{0.95\textwidth}{!}{
  \begin{tabular}{lccc|ccc|ccc|cc|ccccc|cccc}
  \hline
  exp & 1 & 2 & 3 & 4 & 5 & 6 & 7 & 8 & 9 & 10 & 11 & 12 & 13 & 14 & 15 & 16 & 17 & 18 & 19 & 20 \\
  \hline
  g/d & \multicolumn{16}{c|}{generative} & \multicolumn{4}{c}{discriminative} \\
  \hline
  $t_{rgb}$ & 599 & 599 & 599 & 599 & 599 & 599 & 599 & 599 & 599 & 799 & 799 & 991 & 997 & 997 & 997 & 997 & 997 & 991 & 997 & 997 \\
  \hline
  $t_{depth}$ & 599 & 799 & 999 & 699 & 799 & 899 & 999 & 999 & 999 & 899 & 999 & 995 & 998 & 998 & 998 & 998 & 998 & 995 & 599 & 999 \\
  \hline
  $t_{normal}$ & 599 & 799 & 999 & 999 & 999 & 999 & 699 & 799 & 899 & 999 & 899 & 999 & 999 & 999 & 999 & 999 & 999 & 999 & 799 & 999 \\
  \hline
  policy & - & - & - & - & - & - & - & - & - & - & - & - & - & tr & lw & trlw & tr & tr & tr & tr \\
  \hline
  \end{tabular}
  }
\end{table}

\textbf{From Noise Indicator to Categorical Semantic Switch.}
We further investigate the network's sensitivity to specific numerical values of $t$.
Interestingly, we observe a fundamental dichotomy governed by the presence or absence of explicit noise.
In the probabilistic paradigm MUSE-g, where Gaussian noise is explicitly injected as the vanilla diffusion-based dense prediction models \cite{ke2024repurposing, he2024lotus},
 the model remains bound to the physical meaning of $t$ established during pre-training.
Under the premise of following the classic DDPM method and huggingface implementation,
 we observe that higher timestep values (e.g., $t \to 1000$)  yield superior performance. 
This aligns with the diffusion characteristic that high timesteps correspond to a latent space
 dominated by high-level semantic structures rather than low-level noise details, which is crucial for dense geometric reasoning.
The best-performing MUSE-g variant utilizes the three highest timestep values ($t_{rgb}=997$, $t_{depth}=998$, $t_{normal}=999$).

In stark contrast, in the deterministic paradigm MUSE-d, the explicit noise input is removed,
 and the physical interpretation of $t$ as a ``noise-level indicator'' is entirely voided. 
Our ablations reveal that MUSE-d is remarkably robust to the absolute magnitude of $t$.
The timestep transitions purely into a discrete, categorical semantic switch. 
As long as the assigned keys are mutually distinguishable, their specific numerical magnitudes have minimal impact on task decoupling.
This mechanistic transition confirms that in noise-free, one-step diffusion models,
 the native time embedding space is intrinsically rich enough to act as an endogenous multi-task router. 
In this paper, we mainly present MUSE-d with $t_{rgb}=991$, $t_{depth}=995$, $t_{normal}=999$ alongside optimized data-sampling frequencies.

Based on the above findings and the optimal hyperparameter settings,
 we utilize a large batchsize 192 and spent 130 hours to train a MUSE-g and a MUSE-d model that
 aimed for the highest task performance under existing resources. 
The quantitative comparison with other methods are shown in Table \ref{tab:main_results}.
It is worth noting that MUSE achieves performance comparable to recent SOTA multi-task models
 with single step inference, a total parameter size of only 0.95B, and a training set of only 59K.
Detail computing resources comparison are provided in Supp. Materials Sec.C.
Some failure cases of MUSE are shown in Sec.D.
Moreover, experiments on different task numbers and token embedding methods are presented in Supp. Materials Sec.E

\begin{table}[t]
\centering
\caption{
    \textbf{Quantitative Comparison on Dense Prediction Methods.} 
    We evaluate MUSE against SOTA methods across 10 diverse indoor and in-the-wild datasets.
    MUSE achieves highly competitive zero-shot generalization simultaneously on both depth and surface normal estimation.
    Note that for all methods that reported results with different inference steps, we focus only on the performance of one-step inference for fair comparison.
    ``-'' indicates the paper did not report results. The best-performing method for each metric is highlighted in \textbf{bold}.
    We use the same evaluation protocol as Lotus \cite{he2024lotus}.
}
\label{tab:main_results}
\resizebox{\textwidth}{!}{
\begin{tabular}{l cc | cc cc cc cc cc}
\toprule
\multirow{2}{*}{Method} & Gene- & MTL & \multicolumn{2}{c}{NYUv2} & \multicolumn{2}{c}{ScanNet} & \multicolumn{2}{c}{KITTI} & \multicolumn{2}{c}{ETH3D} & \multicolumn{2}{c}{DIODE} \\
\cmidrule(lr){4-5} \cmidrule(lr){6-7} \cmidrule(lr){8-9} \cmidrule(lr){10-11} \cmidrule(lr){12-13}
 & rative & Model & AbsRel$\downarrow$ & $\delta_1\uparrow$ & AbsRel$\downarrow$ & $\delta_1\uparrow$ & AbsRel$\downarrow$ & $\delta_1\uparrow$ & AbsRel$\downarrow$ & $\delta_1\uparrow$ & AbsRel$\downarrow$ & $\delta_1\uparrow$ \\
\midrule
\multicolumn{13}{c}{\cellcolor{gray!15}\textbf{Task 1: Affine-Invariant Depth Estimation}} \\
\midrule
Marigold    & \checkmark        & $\times$   & 6.0 & 95.8 & 6.9 & 94.5 & 10.5 & 90.4 & 7.1 & 95.1 & 31.0 & 77.2 \\
Lotus-g     & \checkmark        & $\times$   & 5.4 & 96.6 & 6.0 & 96.0 & 11.3 & 87.7 & 6.2 & 96.1 & - & - \\
Lotus-d     & $\times$        & $\times$   & 5.3 & 96.7 & 6.0 & 96.3 & 9.3 & 92.8 & 6.8 & 95.3 & - & - \\
Marigoldv1.1 & \checkmark        & $\times$   & 5.9 & 96.1 & 6.6 & 95.3 & 11.0 & 88.8 & 7.0 & 95.5 & 30.4 & 77.3 \\
Lotus-2     & $\times$        & $\times$      & \textbf{4.1} & 97.6 & \textbf{4.2} & \textbf{97.6} & 6.7 & 94.5 & 4.6 & 97.6 & 22.1 & 75.2 \\
Metric3Dv2   & $\times$         & \checkmark   & 4.3 & \textbf{98.1} & - & - & \textbf{4.4} & \textbf{98.2} & 4.2 & 98.3 & \textbf{13.6} & \textbf{89.5} \\
GeoWizard    & \checkmark      & \checkmark   & 5.2 & 96.6 & 6.1 & 95.3 & 9.7 & 92.1 & 6.4 & 96.1 & 29.7  & 79.2 \\
FE2E         & \checkmark      & \checkmark   & 4.1 & 97.7 & 4.4 & 97.5 & 6.6 & 96.0 & \textbf{3.8} & \textbf{98.7} & 22.8 & 81.2 \\
Orchid       & \checkmark      & \checkmark   & 5.7 & 96.9 & 6.3 & 95.8 & 7.7 & 94.4 & 7.3 & 96.9 & - & - \\
Diception    & \checkmark      & \checkmark   & 7.2 & 93.9 & 7.5 & 93.8 & 7.5 & 94.5 & 5.3 & 96.7 & 24.3 & 74.1 \\
\midrule
\rowcolor{blue!5} MUSE-g & \checkmark & \checkmark & 5.8 & 96.1 & 6.9 & 94.3 & 9.8 & 89.9 & 8.8 & 96.2 & 26.3 & 72.5 \\
\rowcolor{blue!5} MUSE-d & $\times$ & \checkmark & 5.1 & 97.1 & 5.8 & 96.2 & 9.2 & 90.8 & 5.9 & 97.1 & 23.8 & 74.0 \\
\midrule
\midrule
\multirow{2}{*}{Method} & Gene- & MTL & \multicolumn{2}{c}{NYUv2} & \multicolumn{2}{c}{ScanNet} & \multicolumn{2}{c}{iBims-1} & \multicolumn{2}{c}{Sintel} & \multicolumn{2}{c}{Oasis} \\
\cmidrule(lr){4-5} \cmidrule(lr){6-7} \cmidrule(lr){8-9} \cmidrule(lr){10-11} \cmidrule(lr){12-13}
 & rative & Model & Mean$\downarrow$ & $11.25^\circ\uparrow$ & Mean$\downarrow$ & $11.25^\circ\uparrow$ & Mean$\downarrow$ & $11.25^\circ\uparrow$ & Mean$\downarrow$ & $11.25^\circ\uparrow$ & Mean$\downarrow$ & $11.25^\circ\uparrow$ \\
\midrule
\multicolumn{13}{c}{\cellcolor{gray!15}\textbf{Task 2: Surface Normal Estimation}} \\
\midrule
Marigold    & \checkmark        & $\times$   & 17.1 & 58.5 & 16.1 & 62.3 & 16.6 & 68.0 & - & - & 23.5 & 28.0 \\
Lotus-g     & \checkmark        & $\times$   & 16.9 & 59.1 & 15.3 & 64.0 & 17.5 & 66.1 & 35.2 & 19.9 & - & - \\
Lotus-d     & $\times$        & $\times$     & 16.8 & 58.2 & 15.3 & 62.9 & 17.7 & 64.9 & 34.6 & 20.5 & - & - \\
Marigoldv1.1 & \checkmark        & $\times$   & 16.4 & 58.9 & 14.9 & 64.3 & 17.2 & 65.6 & - & - & \textbf{23.2} & 28.3 \\
Lotus-2     & $\times$        & $\times$     & 16.9 & 59.0 & 14.2 & 66.8 & 15.4 & 70.4 & \textbf{30.3} & \textbf{27.6} & - & - \\
Metric3Dv2   & $\times$        & \checkmark  & \textbf{13.2} & \textbf{66.2} & -  & - & 19.6 & 69.7 & - & - & 23.4 & 28.5 \\
GeoWizard    & \checkmark      & \checkmark   & 17.0 & 56.5 & 15.4 & 61.6 & \textbf{13.0} & 65.3 & - & - & - & - \\
FE2E         & \checkmark      & \checkmark   & 16.2 & 59.6 & \textbf{13.8} & \textbf{67.2} & 15.1 & \textbf{70.6} & 31.2 & 22.3 & - & - \\
Orchid       & \checkmark      & \checkmark   & 15.2 & 60.6 & 14.2 & 63.8 & 16.3 & 68.1 & 31.7 & 22.6 & - & - \\
Diception    & \checkmark      & \checkmark   & 18.3 & 52.5 & 19.4 & 46.4 & - & - & - & - & - & - \\
\midrule
\rowcolor{blue!5} MUSE-g & \checkmark & \checkmark & 17.1 & 58.0 & 15.7 & 62.7 & 17.4 & 65.8 & 35.9 & 19.0 & 26.2 & 26.6 \\
\rowcolor{blue!5} MUSE-d & $\times$ & \checkmark & 16.6 & 58.6 & 15.4 & 61.3 & 16.7 & 67.3 & 35.2 & 18.6 & 23.4 & \textbf{30.5} \\
\bottomrule
\end{tabular}
}
\end{table}

\subsection{Manifold Decoupling Validation}
\label{sec:manifold}

\begin{figure}[h]
    \centering
    \includegraphics[width=0.98\textwidth]{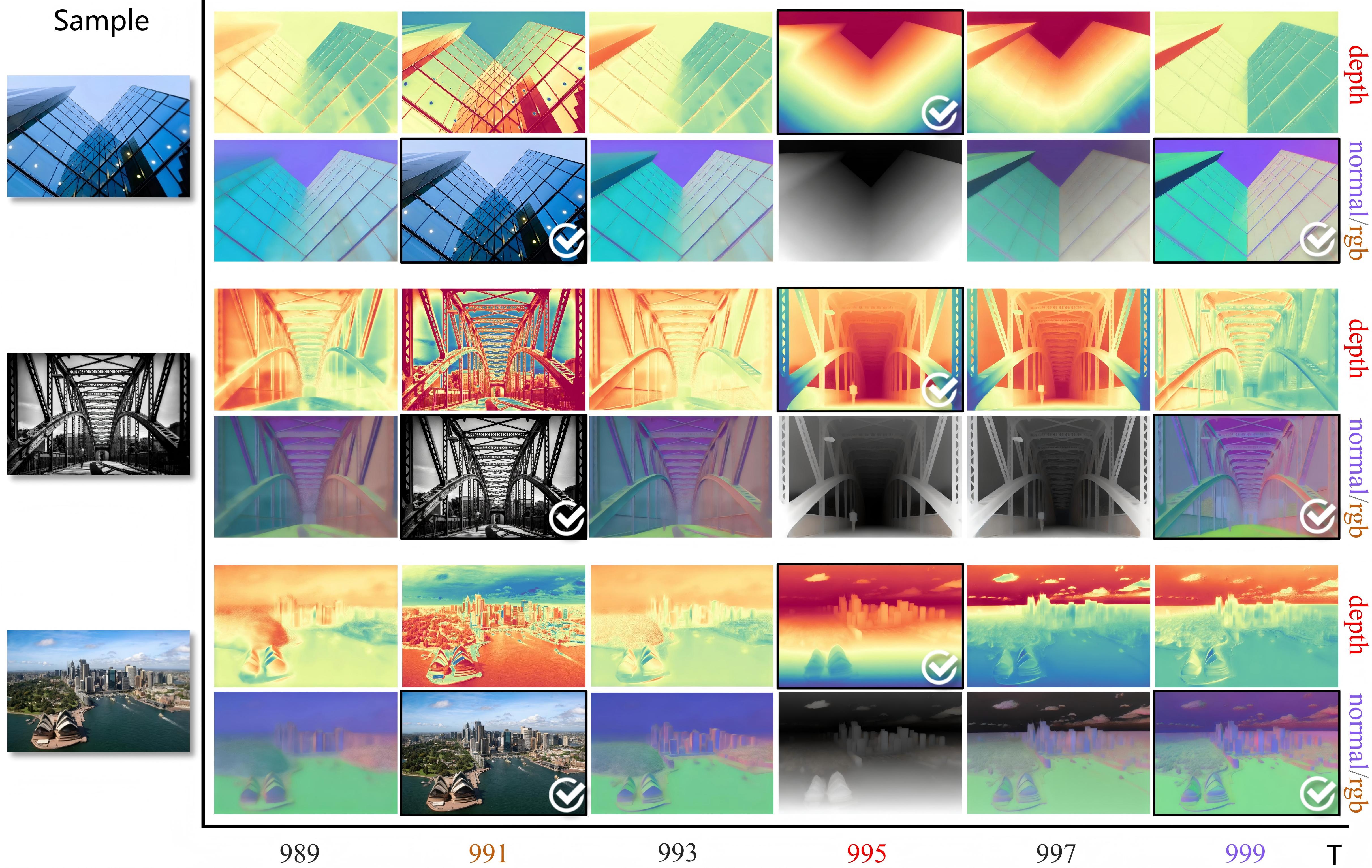}
    \caption{Inference results at different timesteps by exactly the same model.
     The maps marked with a white checkmark are the expected outputs.
     Note that a sample generates only one prediction map at each timestep.
     The depth and normal/rgb images rows are two ways of visualizations, rendered by 1-channel and 3-channel respectively.}
    \label{fig:timestep}
\end{figure}

To validate the Manifold Decoupling claim, we implement two methods to demonstrate that
 the zero-shot latent representations generated for multiple tasks are geometrically and semantically separable.
A MUSE-d model trained with $t_{\text{rgb}}=991,t_{\text{depth}}=995,t_{\text{normal}}=999$ is used as the predictor.

\textbf{Data Manifold of Latent Features.}
We employ Isomap \cite{tenenbaum2000global} to visualize the latent trajectories of diffusion models,
 where Isomap is a nonlinear dimensionality reduction technique that preserves global geodesic distances on a data manifold.
Here we sample 200 images from coco2017 validation set \cite{lin2014microsoft}.
For each image, we perform multiple forward passes using a sequence of different timesteps, extract the diffusion U-Net's final layer output latent. 
Then we apply multidimensional scaling to embed these latent vectors into a 2D plane, preserving their geodesic relationships.

The result is presented in the right of Figure \ref{fig:teaser}.
Latent features at timesteps where no steering function is applied are intertwined.
While at timesteps 995 and 999, the latent vectors clearly segregate into two distinct clusters based on their corresponding task,
 demonstrating a clean separation in the geodesic feature space.
This provides strong evidence that MUSE is not learning a single, entangled multi-task representation.
Instead, it has successfully learned two separate data manifolds $\mathcal{M}_{\text{depth}}$ and $\mathcal{M}_{\text{normal}}$.
The timestep embedding acts as the key, deterministically guiding the U-Net to project the input onto one of these two decoupled geometric structures.
A quantitative metric table of Silhouette Score and Calinski-Harabasz is presented in Supp. Materials Sec.F.

\textbf{Intuitive Visualizations of Prediction Map.}
To more intuitively illustrate the inference results of MUSE under different timesteps,
 Figure \ref{fig:timestep} shows the rendering of the model's prediction maps under two different visualization ways.
MUSE inference returns a 3-channel prediction map basically. 
For depth estimation, the channels are collapsed to a single scalar depth per pixel by taking the channelwise mean, 
and this single-channel map is then normalized and mapped to RGB using colormap.
By contrast, normal estimation outputs are treated intrinsically as 3-channel vector fields.
The predicted map is assumed to encode surface normals in the three channels and is directly converted to an RGB image for visualization.
Naturally, it can observed that when the timestep is 991, MUSE's output is restored to the original image under the normal visualization,
 which is equivalent to performing an RGB image reconstruction task.

These visualization illustrates the internal mechanism of MUSE, providing a concrete validation for our theory.

\subsection{Generality Across Architectures}
To verify that the timestep steering mechanism is not merely an architectural artifact of the U-Net,
 we extend MUSE to the modern Diffusion Transformer (DiT) architecture \cite{peebles2023scalable}.
Specifically, we replace the SDv2 U-Net with the SD3 MMDiT backbone,
 keeping the VAE frozen, and adapt the training paradigm to Conditional Flow Matching (CFM) \cite{lipman2022flow}.
Remarkably, the identical timestep steering protocol successfully decouples the depth and normal estimation tasks within the MMDiT backbone (Table \ref{tab:dit}, Supp. Materials Sec.G).

\begin{table}[h]
\centering
\caption{
    \textbf{Quantitative metrics of MUSE based on DiT.}
    Variants using DiT as the backbone and flow matching as the optimization objective are implemented to verify the generalization of the MUSE mechanism.
    Experiments on 10 datasets show that the timestep steering is effective in Flow Matching, although the metrics are slightly lower than those of the U-Net version due to computational resource limitations.
}
\label{tab:dit}
\resizebox{\textwidth}{!}{
\begin{tabular}{l cc | cc cc cc cc cc}
\toprule
\multirow{2}{*}{Method} & Gene- & MTL & \multicolumn{2}{c}{NYUv2} & \multicolumn{2}{c}{ScanNet} & \multicolumn{2}{c}{KITTI} & \multicolumn{2}{c}{ETH3D} & \multicolumn{2}{c}{DIODE} \\
\cmidrule(lr){4-5} \cmidrule(lr){6-7} \cmidrule(lr){8-9} \cmidrule(lr){10-11} \cmidrule(lr){12-13}
 & rative & Model & AbsRel$\downarrow$ & $\delta_1\uparrow$ & AbsRel$\downarrow$ & $\delta_1\uparrow$ & AbsRel$\downarrow$ & $\delta_1\uparrow$ & AbsRel$\downarrow$ & $\delta_1\uparrow$ & AbsRel$\downarrow$ & $\delta_1\uparrow$ \\
\midrule
\multicolumn{13}{c}{\cellcolor{gray!15}\textbf{Task 1: Affine-Invariant Depth Estimation}} \\
\midrule
MUSE-g-DiT    & \checkmark        & \checkmark   & 8.6 & 92.4 & 10.1 & 88.9 & 11.6 & 86.8 & 9.0 & 92.4 & 23.4 & 71.9 \\
MUSE-d-DiT    & $\times$        & \checkmark   & 8.0 & 93.7 & 8.4 & 92.3 & 10.7 & 87.3 & 10.8 & 92.0 & 24.1 & 72.0 \\
\midrule
\midrule
\multirow{2}{*}{Method} & Gene- & MTL & \multicolumn{2}{c}{NYUv2} & \multicolumn{2}{c}{ScanNet} & \multicolumn{2}{c}{iBims-1} & \multicolumn{2}{c}{Sintel} & \multicolumn{2}{c}{Oasis} \\
\cmidrule(lr){4-5} \cmidrule(lr){6-7} \cmidrule(lr){8-9} \cmidrule(lr){10-11} \cmidrule(lr){12-13}
 & rative & Model & Mean$\downarrow$ & $11.25^\circ\uparrow$ & Mean$\downarrow$ & $11.25^\circ\uparrow$ & Mean$\downarrow$ & $11.25^\circ\uparrow$ & Mean$\downarrow$ & $11.25^\circ\uparrow$ & Mean$\downarrow$ & $11.25^\circ\uparrow$ \\
\midrule
\multicolumn{13}{c}{\cellcolor{gray!15}\textbf{Task 2: Surface Normal Estimation}} \\
\midrule
MUSE-g-DiT    & \checkmark        & \checkmark   & 20.4 & 51.3 & 22.0 & 45.9 & 19.1 & 61.0 & nan & 13.2 & nan & 11.3 \\
MUSE-d-DiT    & $\times$        & \checkmark   & 19.6 & 50.5 & 19.7 & 52.1 & 19.5 & 59.8 & 38.4 & 13.7 & 34.5 & 24.5 \\
\bottomrule
\end{tabular}
}
\end{table}

Due to stringent computational constraints with only one V100-32GB GPU,
 this DiT variant necessitated aggressive optimization trade-offs, including FP16 mixed precision and 8-bit Adam optimizers.
While these quantization effects result in slightly sub-optimal quantitative metrics compared to our fully-tuned U-Net baseline,
 the successful convergence and clear task decoupling serve as a powerful proof-of-concept.
It demonstrates that our proposed manifold decoupling via timestep steering is a fundamental,
 model-agnostic mechanism applicable to both classical diffusion and flow matching paradigms.

\section{Discussion and Conclusion}

This paper addresses the critical challenges of parameter inefficiency and negative transfer
 when adapting pre-trained diffusion models for unified visual perception.
Rather than resorting to complex multi-head architectures, expert networks, heavy-parameter adapters or learnable task tokens,
 we introduce MUSE, a minimalist single-model multi-tasking paradigm.
Our core conclusion establishes that the native, fixed timestep embedding — long considered a mere noise-level indicator — 
 can be repurposed as a powerful, parameter-free semantic switch in one-step generation.
By binding distinct timestep values to specific tasks,
 we successfully guide a single, full-parameter-shared model to perform decoupled geometric estimations.
Extensive experiments validate MUSE's highly competitive performance across 10 diverse datasets,
 and demonstrate that this mechanism naturally generalizes across both classical U-Net and DiT architectures.

\textbf{Limitations on Semantic Segmentation.} While MUSE elegantly resolves inter-task conflicts for continuous geometric targets (e.g., depth and surface normals),
 our explorations reveal distinct boundaries when extending to discrete semantic tasks. 
For instance, we attempted to steer MUSE to concurrently generate NYUv2's 41-class semantic segmentation masks on the Hypersim dataset
 by formulating the discrete class IDs as continuous 3-channel RGB maps optimized via MSE loss.
As anticipated, this resulted in sub-optimal quantitative metrics across the board.
The semantic pixel accuracy dropped significantly, see Supp. Material Sec.H for detailed setups and results.
We hypothesize this degradation is not a failure of the timestep steering mechanism itself,
 but rather a fundamental limitation of standard continuous VAEs in encoding and reconstructing highly discrete,
 non-smooth semantic ID maps without specialized heads or discrete latent spaces.
Resolving this representation gap remains an open challenge for unified diffusion-based perception.

\textbf{Future Directions.} MUSE suggests that the path toward generalist models may
 not solely rely on brute-force scaling or stacking external expert modules. 
Instead, immense value lies in discovering more efficient ways to unlock and
 reshape the representational potential already dormant within a single foundation model. 
In the future, we plan to scale this paradigm to a wider set of semantic and
 intrinsic image decomposition tasks to probe the upper bounds of a single model's task capacity.
Furthermore, while the current selection of orthogonal timesteps is empirical,
 exploring automated mechanisms to discover the most effective orthogonal task vectors
 within the time-embedding space presents a highly promising research direction.

\section*{Acknowledgements}
The research was supported by [Basic Research Center, Innovation Program of Chinese Academy of Agricultural Sciences (CAAS-BRC-SAE-2025-01, CAAS-ASTIP-2026-AII)].

%
%
\bibliographystyle{splncs04}
\bibliography{main}

\clearpage

\appendix
\setcounter{figure}{0}
\setcounter{table}{0}

\section*{\centering Supplementary Materials}
\addcontentsline{toc}{section}{Supplementary Material}


\section{Deconstructing Diffusion Conditioning}
 
Rather than operating solely on noisy latent and the timesteps, the denoising U-Net in SD 
 can accept a rich set of multi-modal inputs to guide the generation. 
When adapted for monocular dense prediction, the U-Net $f_\theta$ can be more accurately represented as a function of a diverse set of conditions:
\begin{equation}
  \label{eq:conditioning}
  \mathbf{z}_{\text{pred}}^{y} = f_\theta(\mathbf{z}^{x}, \epsilon, c_{\text{time}}, c_{\text{text}}, c_{\text{class}})
\end{equation}

(1) The latent representation of the input RGB image $\mathbf{z}^{x}$ provides the essential spatial and semantic context. 
In most methods, it is concatenated with the noisy target latent $z_t$ along the channel dimension.
(2) SD models possess a cross-attention mechanism to align the output with the text prompt embedding $c_{\text{text}}$
 while typically disabled by feeding the model an empty string embedding (``'') in dense prediction.
(3) The timestep embedding $c_{\text{time}}$ informs the network about the current noise level $\sqrt{\bar{\alpha}_t}$ in the input latent $z_t$.
A Sinusoidal Positional Embedding of $t$ is added to the hidden features of the U-Net, allowing the model to learn noise-level-specific denoising operations. 
In multi-step diffusion, this is a dynamic signal that changes at each inference step. 
While in the one-step paradigms, the timestep is fixed to a constant value,
 consequently becoming a static, monolithic signal whose sole function is to command the model to perform a full denoise operation. 
(4) Some multi-task generative models use class labels $c_{\text{class}}$ to explicitly introduce task-specific conditions,
such as textual prompts, one-dimensional vector, or linear projection layer to switch between different functionalities.

In dense prediction, some conditioning channels are repurposed ($\mathbf{z}^{x}$), and others are deactivated ($c_{\text{text}}$, $c_{\text{class}}$),
 thereby leaving the timestep embedding $c_{\text{time}}$ as a vestigial state in the one-step generation paradigm. 

\section{MUSE Steer Policy Ablation Study}
The complete numerical data of evaluation metrics in MUSE steer policy ablation study (Figure.3) is as Table \ref{tab:depthnum} and Table \ref{tab:normalnum}:

\begin{table}[t]
\centering
\caption{Depth estimation metrics on benchmark datasets in Fig.3 left.}
\resizebox{0.9\textwidth}{!}{
\begin{tabular}{lcccccccccc}
\toprule
& \multicolumn{2}{c}{NYUv2} & \multicolumn{2}{c}{KITTI}  & \multicolumn{2}{c}{ETH3D} & \multicolumn{2}{c}{ScanNet} & \multicolumn{2}{c}{DIODE} \\
\cmidrule(lr){2-3} \cmidrule(lr){4-5} \cmidrule(lr){6-7} \cmidrule(lr){8-9} \cmidrule(lr){10-11}
\multirow{-2}{*}{exp} & AbsRel$\downarrow$      & $\delta_1\uparrow$        & AbsRel$\downarrow$                       & $\delta_1\uparrow$                           & AbsRel$\downarrow$      & $\delta_1\uparrow$        & AbsRel$\downarrow$       & $\delta_1\uparrow$         & AbsRel$\downarrow$      & $\delta_1\uparrow$        \\
\midrule
1                       & 27.05        & 50.91      & 37.57                         & 35.01                         & 30.25        & 51.26      & 21.44         & 62.39       & 35.69        & 50.34      \\
2                       & 26.33        & 52.62      & 36.59                         & 36.42                         & 29.11        & 51.58      & 20.89         & 63.92       & 35.33        & 51.07      \\
3                       & 26.05        & 53.49      & 33.86                         & 39.02                         & 34.78        & 52.56      & 20.84         & 64.27       & 37.36        & 52.28      \\
4                       & 21.37        & 63.41      & 26.15                         & 54.50                         & 37.10        & 59.04      & 19.35         & 67.97       & 37.06        & 53.91      \\
5                       & 18.29        & 70.32      & 24.67                         & 58.16                         & 37.97        & 67.01      & 17.13         & 73.32       & 37.03        & 58.82      \\
6                       & 16.73        & 74.59      & 15.60                         & 79.20                         & 35.70        & 79.04      & 16.94         & 73.73       & 33.77        & 61.06      \\
7                       & 10.70        & 88.26      & 10.29                         & 88.79                         & 24.58        & 91.01      & 12.97         & 83.40       & 30.74        & 66.39      \\
8                       & 9.69         & 90.27      & 10.03                         & 88.82                         & 24.24        & 91.14      & 11.91         & 85.43       & 29.83        & 67.45      \\
9                       & 9.67         & 89.45      & 10.64                         & 88.53                         & 25.85        & 91.83      & 11.42         & 85.59       & 28.89        & 67.73      \\
10                      & 14.97        & 78.63      & 18.39                         & 73.01                         & 31.92        & 76.77      & 14.94         & 78.62       & 34.84        & 62.15      \\
11                      & 9.05         & 91.55      & 10.43                         & 88.78                         & 25.71        & 91.49      & 10.92         & 87.36       & 29.22        & 67.64      \\
12                      & 14.51        & 80.43      & 15.89                         & 78.22                         & 29.79        & 81.37      & 15.70         & 77.16       & 33.62        & 63.13      \\
13                      & 11.76        & 86.51      & 11.69                         & 86.35                         & 34.41        & 88.48      & 13.33         & 82.49       & 32.44        & 65.79      \\
14                      & 8.95         & 91.57      & 10.93                         & 88.19                         & 20.24        & 91.74      & 10.61         & 88.04       & 26.87        & 69.38      \\
15                      & 11.48        & 86.65      & 12.59                         & 85.17                         & 28.36        & 85.89      & 12.59         & 83.94       & 28.84        & 66.93      \\
16                      & 11.16        & 87.62      & 11.64                         & 86.64                         & 27.88        & 90.12      & 12.46         & 84.31       & 30.52        & 67.29      \\
17                      & 6.31         & 95.86      & 9.81                          & 89.68                         & 7.37         & 95.97      & 7.24          & 94.44       & 23.34        & 73.52      \\
18                      & 6.40         & 95.76      & 10.07 & 89.51 & 7.58         & 94.92      & 7.17          & 94.29       & 24.27        & 72.42      \\
19                      & 6.48         & 95.76      & 9.782                         & 89.66                         & 7.53         & 95.81      & 7.16          & 94.82       & 23.35        & 73.35      \\
20                      & 10.70        & 88.56      & 16.85 & 75.54 & 27.02        & 84.90      & 11.31         & 87.38       & 29.38        & 70.08      \\
\bottomrule
\end{tabular}
}
\label{tab:depthnum}
\end{table}

\begin{table}[t]
\centering
\caption{Normal estimation metrics on benchmark datasets in Fig.3 right.}
\resizebox{0.9\textwidth}{!}{
\begin{tabular}{lcccccccccc}
\toprule
& \multicolumn{2}{c}{NYUv2}  & \multicolumn{2}{c}{ScanNet} & \multicolumn{2}{c}{iBims-1}   & \multicolumn{2}{c}{Sintel}   & \multicolumn{2}{c}{Oasis} \\ 
\cmidrule(lr){2-3} \cmidrule(lr){4-5} \cmidrule(lr){6-7} \cmidrule(lr){8-9} \cmidrule(lr){10-11}
\multirow{-2}{*}{exp} & Mean$\downarrow$                         & $11.25^{\circ}\uparrow$ & Mean$\downarrow$                         & $11.25^{\circ}\uparrow$                       & Mean$\downarrow$                         & $11.25^{\circ}\uparrow$                       & Mean$\downarrow$                         & $11.25^{\circ}\uparrow$                       & Mean$\downarrow$       & $11.25^{\circ}\uparrow$     \\
\midrule
1                       & 112.50                        & 0.30    & 120.20                        & 0.16                          & 98.97                         & 0.48                          & nan                           & 0.52                          & 113.50      & 0.38        \\
2                       & 60.27                         & 9.34    & 78.17                         & 3.34                          & 58.68                         & 13.21                         & nan                           & 3.17                          & 79.45       & 3.72        \\
3                       & 24.16                         & 43.71   & 21.76                         & 44.27                         & 23.84                         & 52.16                         & 48.55                         & 9.57                          & 35.41       & 14.66       \\
4                       & 19.83                         & 51.68   & 20.64                         & 49.08                         & 19.74                         & 58.37                         & 43.61                         & 11.46                         & 33.10       & 15.59       \\
5                       & 21.00                         & 50.33   & 20.71                         & 47.78                         & 19.93                         & 57.70                         & 39.89                         & 12.88                         & 28.90       & 19.39       \\
6                       & 20.56                         & 48.08   & 22.35                         & 47.26                         & 20.15                         & 56.43                         & 38.75                         & 14.54                         & 27.14       & 20.78       \\
7                       & 69.60                         & 3.445   & 84.47                         & 2.21                          & 62.20                         & 5.43                          & nan                           & 1.04                          & 98.05       & 1.79        \\
8                       & 41.04                         & 14.48   & 57.03                         & 8.73                          & 40.53                         & 19.61                         & 69.30                         & 2.94                          & 67.62       & 5.81        \\
9                       & 25.79                         & 32.84   & 30.31                         & 24.53                         & 25.51                         & 40.22                         & 60.68                         & 9.16                          & 49.48       & 15.36       \\
10                      & 19.50                         & 51.56   & 20.32                         & 47.94                         & 18.30                         & 61.49                         & 41.43                         & 14.31                         & 30.79       & 21.35       \\
11                      & 27.60                         & 38.61   & 35.79                         & 27.08                         & 27.33                         & 45.51                         & 62.11                         & 9.14                          & 48.20       & 14.95       \\
12                      & 19.85                         & 53.02   & 19.75                         & 52.27                         & 18.64                         & 59.94                         & 46.40                         & 12.25                         & 32.63       & 18.72       \\
13                      & 19.74                         & 51.33   & 19.98                         & 47.36                         & 19.67                         & 56.46                         & 40.28                         & 13.68                         & 27.26       & 23.58       \\
14                      & 20.13                         & 51.60   & 22.68                         & 48.63                         & 19.67                         & 60.92                         & 50.11                         & 13.80                         & 40.31       & 18.50       \\
15                      & 20.12                         & 53.42   & 22.53                         & 50.47                         & 18.61                         & 60.95                         & 44.98                         & 12.37                         & 30.35       & 22.98       \\
16                      & 19.36                         & 53.88   & 19.82                         & 49.82                         & 18.85                         & 59.31                         & nan                           & 13.82                         & 34.52       & 20.63       \\
17                      & 17.57                         & 56.34   & 17.07                         & 55.77                         & 17.60                         & 64.09                         & 37.28                         & 13.38                         & 25.92       & 26.86       \\
18                      & 19.28 & 53.36   & 17.36 & 56.51 & 17.93 & 63.03 & 36.87 & 14.43 & 26.24       & 24.44       \\
19                      & 19.07                         & 52.04   & 17.73                         & 53.01                         & 17.97                         & 63.59                         & 36.60                         & 13.38                         & 25.57       & 26.28       \\
20                      & 78.89 & 2.64    & 63.98 & 6.16  & 17.30 & 3.51  & 80.15 & 2.01  & 75.94       & 2.86        \\
\bottomrule
\end{tabular}
}
\label{tab:normalnum}
\end{table}

\section{Comparison of Computing Resources}
Table \ref{tab:resource} shows that recent SOTA methods rely on massive computing, large-scale datasets and likely more inference steps.
In contrast, MUSE achieves highly competitive joint prediction using zero extra parameters, one step, and minimal resources.

\begin{table}[h]
\centering
\caption{Methods computing resource comparison.
 Number of parameters refers to the total model parameters required in GPU memory during inference (estimated).
 ``-'' indicates that this information was not reported in the original paper.}
\label{tab:resource}
\resizebox{0.7\textwidth}{!}{
\begin{tabular}{l l l l l l l l}
\toprule
Method & Backbone & Parameter & Data & Training GPU Days \\
\hline
Marigold & SDv2 U-Net & 0.95B  & 74K & 1$\times$4090 (24G)$\times$2.5d  \\
Lotus & SDv2 U-Net & 0.95B  & 59K & 8$\times$A800 (80G)$\times$8h  \\
Lotus-2 & FLUX+LoRA & 12B  & 59K & 8$\times$H100 (80G)$\times$-- \\
GeoWizard & SDv2 U-Net & 0.95B  & 190K & 8$\times$A100 (40G)$\times$2d \\
DICEPTION & DiT+LoRA & 8B  & 1.8M & 4$\times$H800 (80G)$\times$24d \\
Orchid & DiT & 2.5B  & 2.5M & 16$\times$A100 (40G)$\times$7d \\
FE2E & FLUX+LoRA & 12B  & 71K & 1$\times$H20 (96G)$\times$1.5d \\
Metric3Dv2 & ViT-g + DPT & 1.2B & 16M & 48$\times$A100 (40G)$\times$-- \\
\hline
MUSE & SDv2 U-Net & 0.95B  & 59K & 1$\times$V100 (32G)$\times$5.5d \\
\bottomrule
\end{tabular}%
}
\label{tab:resource}
\end{table}

\section{Failure Cases}

\begin{figure}[tb]
  \centering
  \includegraphics[height=6.5cm]{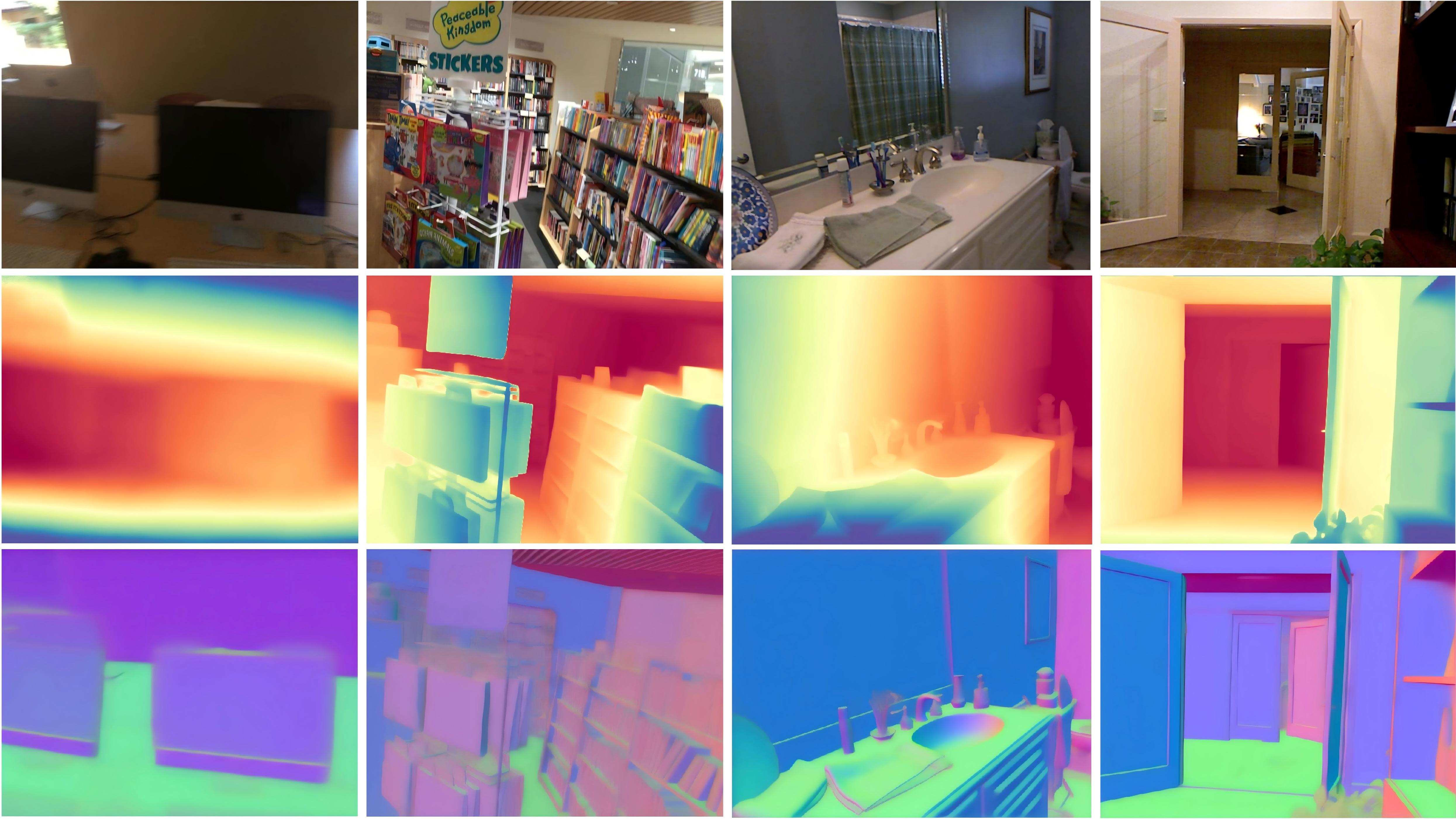}
  \caption{Some failure cases.
  }
  \label{fig:badcase}
\end{figure}

As Figure \ref{fig:badcase} shows, in general, MUSE typically fails on scenes with severe blur, transparent surfaces, and fragmented micro-textures.
Theoretically, a one-step estimator struggles to correct geometric optical illusions,
 when the RGB texture cues extracted by the VAE are corrupted.

\section{Experiments of Task Number and Timestep Embedding Form}

We trained strict double\&single-task MUSE models under identical settings (especially the trainning step).
In Table \ref{tab:tasknum}, the results empirically falsify the presence of negative transfer.
This quantitative parity validates that our timestep steering mechanism successfully mitigates gradient conflicts.

We ablated our baseline against Fourier timestep embedding and standard learnable task tokens (nn.Embedding).
Table \ref{tab:tasknum} proves that unlocking the pre-trained time-embedding space is not merely an equivalent alternative,
 but is natively superior to other fixed embedding methods or introducing routing parameters.

\begin{table}[h]
\centering
\caption{Experiments on the task number and token embedding methods.}
\resizebox{0.8\linewidth}{!}{%
\begin{tabular}{lccccccccc}
\toprule
                        &                                                                        &                                                                        &                          &                        &                        & \multicolumn{2}{c|}{NYUv2 (Depth)} & \multicolumn{2}{c}{iBims-1 (Normal)}                                   \\ \cline{7-10} 
\multirow{-2}{*}{model} & \multirow{-2}{*}{\begin{tabular}[c]{@{}c@{}}train\\ step\end{tabular}} & \multirow{-2}{*}{\begin{tabular}[c]{@{}c@{}}task\\ ratio\end{tabular}} & \multirow{-2}{*}{$t_{rgb}$} & \multirow{-2}{*}{$t_d$} & \multirow{-2}{*}{$t_n$} & AbsRel$\downarrow$      & \multicolumn{1}{c|}{$\delta$1$\uparrow$}        & Mean$\downarrow$                          & 11.25°$\uparrow$                       \\
\hline
depth                   & 5k                                                                     & /                                                                      & /                        & 999                    & /                      & 10.70        & 88.37      & \multicolumn{2}{c}{/}                                         \\
normal                  & 5k                                                                     & /                                                                      & /                        & /                      & 999                    & \multicolumn{2}{c}{/}     & 20.01                         & 57.85                         \\
d+n                     & 10k                                                                    & 1:1                                                                    & /                        & 998                    & 999                    & 11.07        & 87.68      & 19.66                         & 60.97                         \\
fourier                 & 12k                                                                    & 2:5:5                                                                  & 997                      & 998                    & 999                    & 10.02        & 89.44      & 19.48                         & 58.07                         \\
learnable               & 12k                                                                    & 2:5:5                                                                  & 999                      & 999                    & 999                    & 12.76        & 83.63      & 20.83                         & 58.17                         \\
\hline
MUSE                    & 12k                                                                    & 2:5:5                                                                  & 997                      & 998                    & 999                    & 8.95         & 91.57      & 19.67 & 60.92 \\
\bottomrule
\end{tabular}
}
\label{tab:tasknum}
\end{table}

\section{Quantitative Metrics of Manofold Decoupling}

We computed the Silhouette Score (SS) and Calinski-Harabasz (CH) Index for the UNet middle\&final layers' latent features.
Silhouette Score measures cluster compactness and separation by comparing intra-cluster cohesion to inter-cluster distance, ranging from -1 (poor) to +1 (well-separated).
Calinski-Harabasz Index evaluates clustering quality as the ratio of between-cluster dispersion to within-cluster dispersion; higher values indicate better-defined clusters.
Compared with unsteered timesteps, [995,999] yields a dramatically improved SS and CH (Table \ref{tab:SSCH}).

\begin{table}[h]
\centering
\caption{Silhouette Score and Calinski-Harabasz index of features.}
\label{tab:SSCH}
\resizebox{0.6\linewidth}{!}{%
\begin{tabular}{ccccc}
\toprule
Timesteps             & SS@mid$\uparrow$ & CH@mid$\uparrow$ & SS@last$\uparrow$ & CH@last$\uparrow$ \\ \hline
{[}599,699,799,899{]} & 0.0001 & 6.01   & -0.0166 & 0.89   \\
{[}995,999{]}         & 0.0487 & 22.10  & 0.1572  & 80.27   \\
\bottomrule
\end{tabular}
}
\label{tab:SSCH}
\end{table}

\section{MUSE Generality on DiT Architecture}

To evaluate the architectural generalization of MUSE formulation,
 we extended the U-Net (\textit{UNet2DConditionModel}) backbone to a Multimodal Diffusion Transformer (MMDiT), utilizing the \textit{SD3Transformer2DModel}.
This transition from the U-Net to the DiT architecture necessitated several fundamental modifications to both the network structure and the generative pipeline.

First, the optimization framework was conceptually shifted from the DDPM $x_0$-prediction formulation to Conditional Flow Matching (CFM).
The scheduler was replaced accordingly, and the forward noising process was reformulated to construct mixed states via linear interpolation between the clean data and Gaussian noise based on normalized timesteps.
Notably, the MMDiT is trained to predict the original sample following $x_0$-prediction objective. 
Second, substantial architectural adaptations were implemented at the input-output interfaces.
Aligning with the Stable Diffusion 3 framework, we incorporated a 16-channel VAE latent space. 
To facilitate image-to-image translation, the patch embedding layer of the MMDiT was explicitly modified to expand its input capacity from 16 to 32 channels.
This enables the direct channel-wise concatenation of the 16-channel encoded RGB condition and the 16-channel noisy target latent.
Crucially, despite these structural and functional replacements, the core timestep steering paradigm was strictly preserved.
The model continues to optimize multiple dense prediction tasks — specifically depth estimation, surface normal estimation,
 and RGB reconstruction by mapping each specific task to a distinct, dynamically assigned timestep.

This experiment successfully demonstrates that the MUSE mechanism generalizes robustly across fundamentally different generative backbones.

\section{MUSE Expriments of Semantic Segmentation}

Building upon the MUSE framework, we augmented the dense prediction paradigm to simultaneously optimize four distinct tasks:
 RGB reconstruction, depth estimation, surface normal prediction, and semantic segmentation.
The core architecture retains the unified U-Net backbone, avoiding any task-specific specialized prediction heads.
The semantic segmentation task is seamlessly integrated via the assignment of another unique, dedicated diffusion timestep.
During training, the gradients for all four tasks are derived from a unified latent Mean Squared Error (MSE) objective.

To align discrete, categorical semantic labels with our continuous generative formulation,
 we fundamentally recast semantic segmentation as an image-to-image pseudo-color regression task.
Utilizing the Semantic part of Hypersim dataset, we established a deterministic mapping dictionary
 that translates the original 41-class semantic ID maps (along with an invalid/void class) into 3-channel RGB images.
These synthesized semantic pseudo-color images are subsequently normalized to the [-1,1] continuous range
 and compressed into the model's native working space via the VAE.
Consequently, the model processes semantic constraints totally identical to depth and normal signals,
 fully reusing the existing cross-attention and convolutional encoding pathways.

During the evaluation phase, the generative pipeline samples the semantic timesteps
 and decodes the latent representations back into spatial RGB semantic maps.
To reinstate the discrete class assignments, an inverse mapping procedure is executed:
 the predicted continuous RGB arrays are transformed back to class IDs by
 computing the minimal pixel-wise L2 distance against the predefined color lookup table.
The quantitative assessment natively omits predefined void regions (ID=-1).
The final task performance is evaluated on the validation split of Hypersim Semantic dataset
 utilizing pixel accuracy and mean Intersection over Union (mIoU).

\begin{table}[h]
\centering
\caption{
    Quantitative metrics of MUSE extended to semantic segmentation task.
    The timestep assigned to semantic segmentation is 996, with $t_{rgb}=997$, $t_{depth}=998$, $t_{normal}=999$.
    The task ratio and loss weight here is 1:1:1:1, and the model was trained for 16K iterations due to the increase in the number of tasks.
}
\label{tab:dit}
\resizebox{\textwidth}{!}{
\begin{tabular}{l cc | cc cc cc cc cc}
\toprule
\multirow{2}{*}{Method} & Gene- & MTL & \multicolumn{2}{c}{NYUv2} & \multicolumn{2}{c}{ScanNet} & \multicolumn{2}{c}{KITTI} & \multicolumn{2}{c}{ETH3D} & \multicolumn{2}{c}{DIODE} \\
\cmidrule(lr){4-5} \cmidrule(lr){6-7} \cmidrule(lr){8-9} \cmidrule(lr){10-11} \cmidrule(lr){12-13}
 & rative & Model & AbsRel$\downarrow$ & $\delta_1\uparrow$ & AbsRel$\downarrow$ & $\delta_1\uparrow$ & AbsRel$\downarrow$ & $\delta_1\uparrow$ & AbsRel$\downarrow$ & $\delta_1\uparrow$ & AbsRel$\downarrow$ & $\delta_1\uparrow$ \\
\midrule
\multicolumn{13}{c}{\cellcolor{gray!15}\textbf{Task 1: Affine-Invariant Depth Estimation}} \\
\midrule
MUSE-g-Seg    & \checkmark        & \checkmark   & 10.8 & 88.1 & 12.8 & 83.3 & 13.2 & 83.3 & 49.4 & 82.3 & 31.8 & 66.2 \\
\midrule
\midrule
\multirow{2}{*}{Method} & Gene- & MTL & \multicolumn{2}{c}{NYUv2} & \multicolumn{2}{c}{ScanNet} & \multicolumn{2}{c}{iBims-1} & \multicolumn{2}{c}{Sintel} & \multicolumn{2}{c}{Oasis} \\
\cmidrule(lr){4-5} \cmidrule(lr){6-7} \cmidrule(lr){8-9} \cmidrule(lr){10-11} \cmidrule(lr){12-13}
 & rative & Model & Mean$\downarrow$ & $11.25^\circ\uparrow$ & Mean$\downarrow$ & $11.25^\circ\uparrow$ & Mean$\downarrow$ & $11.25^\circ\uparrow$ & Mean$\downarrow$ & $11.25^\circ\uparrow$ & Mean$\downarrow$ & $11.25^\circ\uparrow$ \\
\midrule
\multicolumn{13}{c}{\cellcolor{gray!15}\textbf{Task 2: Surface Normal Estimation}} \\
\midrule
MUSE-g-Seg    & \checkmark        & \checkmark   & 19.5 & 49.8 & 21.1 & 41.6 & 20.3 & 53.6 & 46.0 & 11.3 & 27.9 & 21.8 \\
\midrule
\midrule
\multirow{2}{*}{Method} & Gene- & MTL & \multicolumn{2}{c}{Hypersim}  \\
\cmidrule(lr){4-5} 
 & rative & Model & mIoU$\uparrow$ & Pixel Acc$\uparrow$  \\
\midrule
\multicolumn{13}{c}{\cellcolor{gray!15}\textbf{Task 3: Semantic Segmentation}} \\
\midrule
MUSE-g-Seg    & \checkmark        & \checkmark   & 15.4 & 46.2  \\
\bottomrule
\end{tabular}
}
\end{table}

\end{document}